\documentclass[conference]{IEEEtran}
\IEEEoverridecommandlockouts
\makeatletter
\newcommand{\onedot}{.}
\def\@onedot{\ifx\@let@token.\else.\null\fi\xspace}

\def\etal{\emph{et al}\onedot}
\makeatother

\usepackage{cite}
\usepackage{amsmath,amssymb,amsfonts}
\usepackage{nccmath}
\usepackage{algorithmic}
\usepackage{graphicx}
\usepackage{textcomp}
\usepackage{xcolor}
\def\BibTeX{{\rm B\kern-.05em{\sc i\kern-.025em b}\kern-.08em
    T\kern-.1667em\lower.7ex\hbox{E}\kern-.125emX}}

\usepackage{latexsym}
\usepackage{comment}
\usepackage{multirow}
\usepackage{enumitem}
\usepackage{siunitx}
\setlist{nolistsep} 
\usepackage{adjustbox}
\usepackage{booktabs}
\usepackage{flushend}

\usepackage{url}
\usepackage{xcolor}
\definecolor{newcolor}{rgb}{.8,.349,.1}

\usepackage[pagebackref=true,breaklinks=true,colorlinks,bookmarks=false]{hyperref}
\usepackage{nicefrac}

\usepackage[capitalize]{cleveref}
\crefname{section}{section}{sections}
\Crefname{section}{Section}{Sections}
\crefname{table}{Table}{Tables}
\Crefname{table}{Table}{Tables}
\crefname{figure}{Fig.}{Fig.}
\Crefname{figure}{Fig.}{Fig.}

\begin{document}
\title{Can Shape-Infused Joint Embeddings Improve Image-Conditioned 3D Diffusion?}

\author{
Cristian Sbrolli \quad Paolo Cudrano \quad Matteo Matteucci\\
Department of Electronics Information and Bioengineering\\
Politecnico di Milano, Italy \\
{\{name.surname\}@.polimi.it}
}
\maketitle

\begin{abstract}
Recent advancements in deep generative models, particularly with the application of CLIP (Contrastive Language–Image Pre-training) to Denoising Diffusion Probabilistic Models (DDPMs), have demonstrated remarkable effectiveness in text-to-image generation. The well-structured embedding space of CLIP has also been extended to image-to-shape generation with DDPMs, yielding notable results. Despite these successes, some fundamental questions arise: Does CLIP ensure the best results in shape generation from images? Can we leverage conditioning to bring explicit 3D knowledge into the generative process and obtain better quality? This study introduces CISP (Contrastive Image-Shape Pre-training), designed to enhance 3D shape synthesis guided by 2D images. CISP aims to enrich the CLIP framework by aligning 2D images with 3D shapes in a shared embedding space, specifically capturing 3D characteristics potentially overlooked by CLIP's text-image focus. Our comprehensive analysis assesses CISP's guidance performance against CLIP-guided models, focusing on generation quality, diversity, and coherence of the produced shapes with the conditioning image. We find that, while matching CLIP in generation quality and diversity, CISP substantially improves coherence with input images, underscoring the value of incorporating 3D knowledge into generative models. These findings suggest a promising direction for advancing the synthesis of 3D visual content by integrating multimodal systems with 3D representations.
\end{abstract}


\section{Introduction}
\label{sec:introduction}
Recent years have witnessed remarkable advancements in deep generative models, especially in image generation. Denoising Diffusion Probabilistic Models (DDPMs) played a central role in this progress, outperforming previous methods such as variational autoencoders (VAEs) and generative adversarial networks (GANs) not only in unconditional image synthesis~\cite{diffBeatGANs}, but also in text-to-image synthesis. Works such as GLIDE~\cite{GLIDE}, DALLE\nobreakdash-2~\cite{DALLE2}, Stable Diffusion~\cite{stablediffusion}, and Imagen~\cite{Imagen} showed, indeed, how we can effectively condition the generation of images by text prompts.

Behind the effectiveness of many conditioned DDPMs is the guidance provided by joint embedding models, such as CLIP~\cite{CLIP} in the case of DALLE\nobreakdash-2~\cite{DALLE2}. The strength of these models lies in mapping multimodal concepts in the same well-structured embedding space, thus effectively aligning the modalities together.
This mechanism aids diffusion models in generating content in one modality that is highly coherent with a guiding source in any other modality. In the case of DALLE\nobreakdash-2 with CLIP, guidance is provided through textual prompts.

The undeniable success of diffusion on 2D images led research to focus also on other modalities.
In particular, interest has sparked in the generation of 3D shapes guided by 2D images. Image-to-shape generation is beneficial to fields such as VR and AR, cultural heritage~\cite{culturalheritage}, medical imaging and diagnosis~\cite{medical}, and industrial design and manufacturing~\cite{industrial}. Inspired by text-to-image generation, LION~\cite{lion} showed that it is possible to use the CLIP embedding of a 2D image to condition a diffusion model in the generation of a similar 3D shape, achieving state-of-the-art performance in quality and diversity.
Yet, this result also begs the question of whether aligning text to 2D information
is all that is needed to produce image-guided shapes.
\begin{figure}[t]
  \centering
   \includegraphics[width=1\linewidth]{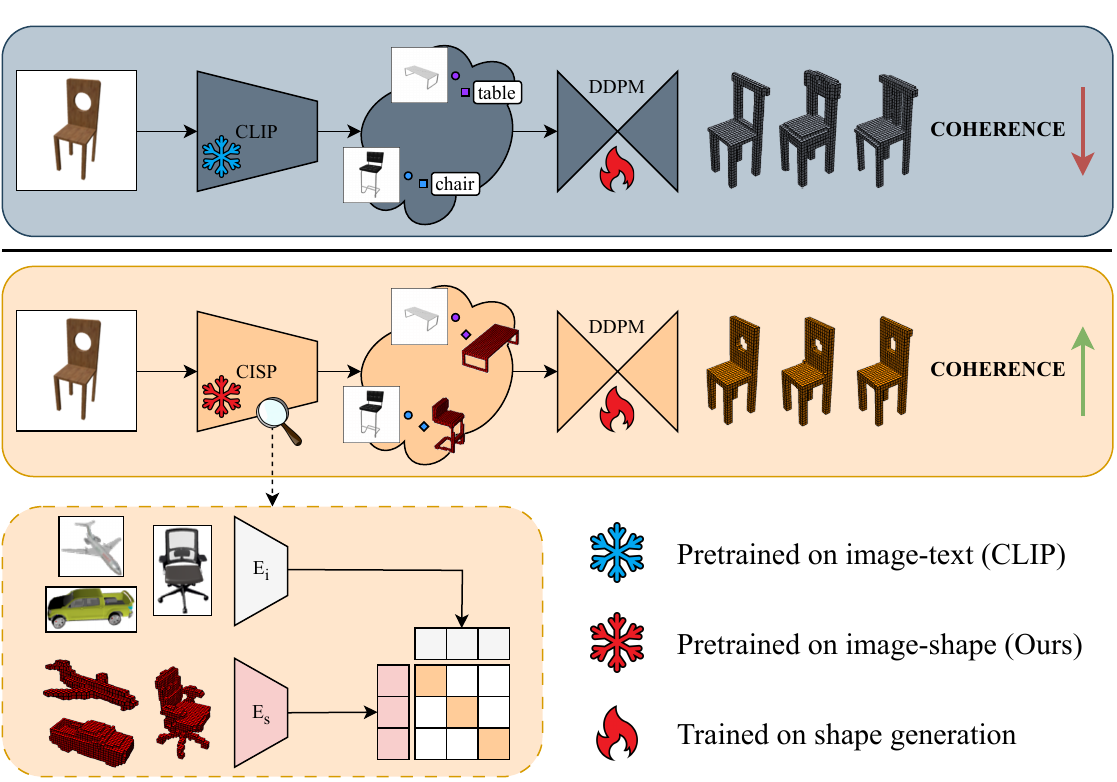}
   \caption{Overview of our study. Our analysis explores the impact of employing image-shape embedding spaces versus text-image embedding spaces for generating 3D shapes from images using DDPMs. The results indicate that both models achieve satisfactory quality and diversity. However, the use of 3D-aware embeddings demonstrates enhanced alignment and consistency between the generated 3D shapes and the original conditioning images.}
\end{figure}
In fact, during training, CLIP is only provided with two-dimensional visual features and does not have access to any structural information about the shapes behind the images it observes. Besides, obstruction and visual ambiguities in 2D images are very common, and injecting only 2D information in the guiding process might lead to diverse shapes collapsing into similar embeddings.

Motivated by this question, with this work we set out to assess the impact of adding 3D information in the guiding process. We propose a joint-embedding model, named CISP (Contrastive Image-Shape Pre-training), analogous to CLIP but designed to align 2D images and 3D shapes in a joint embedding space. As CISP is trained using a contrastive loss, we expect the structure itself of the embedding space to reflect 3D characteristics that might be invisible or cluttered in the mere 2D representations of the same objects.
We assess the influence of CISP embeddings compared to text-image CLIP embeddings when employed as guidance for a shape diffusion model. To achieve this goal, we establish a generation pipeline for image-guided 3D diffusion that allows conditioning on either CISP and CLIP embeddings. We quantitatively compare their generation performance in terms of generation quality and diversity, as well as the coherence of the resulting shapes to the query image. We further compare the properties and structure of the two embedding spaces through interpolations. Finally, we assess the OOD capabilities of both models and perform ablation on the proposed architecture.

This study contributes to the ongoing dialogue on generative models,
paving the way for innovative three-dimensional visual content synthesis.
As our small-scale experiments show promising results, we argue that investing in training a large-scale multimodal system including 3D representations would be highly beneficial to advance the generation capabilities of current models to the 3D world.

In summary, the key contributions of this work are:
\begin{itemize}
    \item We present CISP, a model exploiting contrastive pre-training to learn joint image-shape embeddings.
    \item We show that a CISP-conditioned diffusion model generates shapes with higher coherence to the guiding images than a CLIP-conditioned model, while maintaining similar generation quality and diversity. This phenomenon occurs despite the large-scale nature of CLIP compared to CISP.
    \item We investigate the impact of shape-aware embeddings by studying the regularity of the CISP embedding space with manifold interpolations and out-of-distribution sampling.
\end{itemize}

\section{Related Work}
\label{sec:related_work}
\subsection{3D Generation}
Different approaches have been proposed to obtain generative models capable of synthesizing 3D objects. Early advances were marked by the introduction of 3D Variational Autoencoders such as 3D-VAE~\cite{3D-VAE}, which proposed a voxel-based architecture for 3D generation.   Building on the success in two-dimensional spaces, Generative Adversarial Networks (GAN)~\cite{GANs} were adapted for creating 3D voxelized shapes, a notable example being 3D-GAN. Despite achieving impressive results, GANs often face challenges related to training stability, primarily due to the adversarial nature of their generator and discriminator networks. This issue spurred interest in alternative generative approaches, including flow-based models~\cite{normalizingFlows}, Energy-Based Models (EBMs)~\cite{energybasedtutorial}, and Denoising Diffusion Probabilistic Models (DDPMs)~\cite{ddpmOriginal, ddpm}. Flow-based models, such as DPF-Net~\cite{dpfnet}, manipulate probability distributions to generate samples through variable transformations. EBMs, exemplified by 3D DescriptorNet~\cite{3DDesNet}, optimize energy functions over observed variables and generate new data via Langevin Dynamics~\cite{implicitLangevin, learningMCMC}. In contrast, Denoising Diffusion Probabilistic Models (DDPMs) denoise a progressively noised input sample, a process achieved through a forward diffusion mechanism. Once trained, the model is utilized to reverse this forward diffusion process, thereby enabling the generation of samples from mere noise. They have proven to be extraordinarily effective in text-to-image generation works such as DALLE\nobreakdash-2~\cite{DALLE2} and Stable Diffusion~\cite{stablediffusion}. Following their success in two-dimensional generation, the extension of DDPMs to three-dimensional generation became a logical progression.

Initial explorations in 3D Denoising Diffusion Probabilistic Models (DDPMs), such as PVD~\cite{PVD}, have demonstrated their effectiveness in generating unconditional shapes, surpassing earlier generative techniques. PVD employs Point\nobreakdash-Voxel~CNN~\cite{pvCNN} to generate point cloud shapes, while also reporting challenges in training voxel-based DDPMs. Due to its unconditional approach, PVD requires distinct models and training processes for each new shape category. In contrast, Luo and Hu~\cite{luo2021diffusion} introduce a point cloud DDPM conditioned on shape latents derived from a point cloud autoencoder. This advancement allows a single model to produce several object categories, showcasing the versatility of conditioned DDPMs. Hui \etal~\cite{wavelet} have applied diffusion processes to SDFs wavelet coefficients, generating coarse volumes and utilizing a refiner network for detailing. Leveraging latent diffusion, LION~\cite{lion} demonstrates image-conditioning of a 3D generation model using CLIP~\cite{CLIP} text or image embeddings. However, although well-structured, CLIP embeddings inherently lack 3D features, and their use in 3D generation might lead to loss of fine-grained structural shape properties and details that are not captured by images or text. We aim to assess these limitations by employing joint image-shape embeddings.

\begin{figure*}[t]
  \centering
   \includegraphics[width=0.9\linewidth]{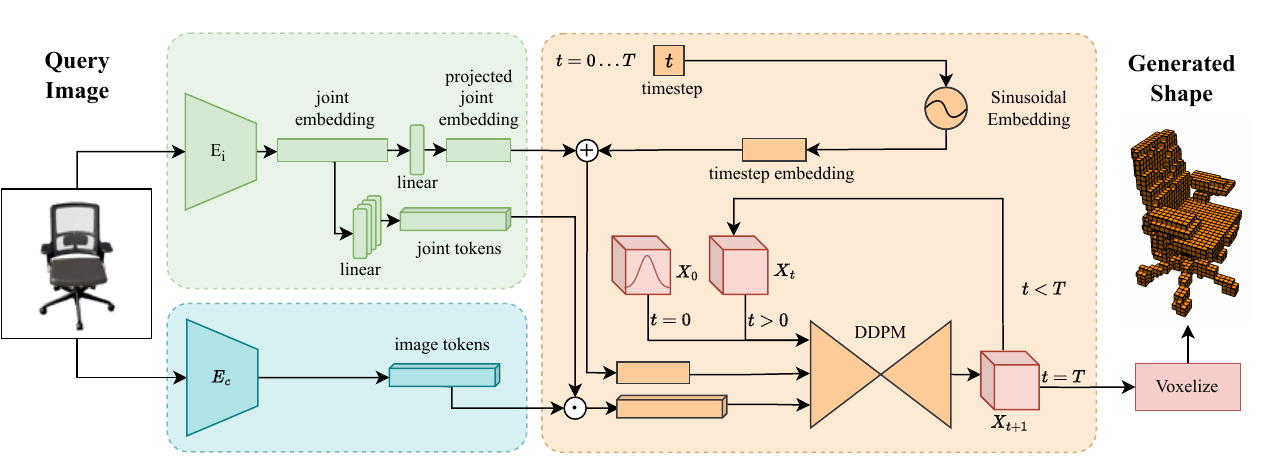}
   \caption{
   Our image-conditioned 3D generation pipeline. The query image is processed via a pre-trained image encoder $E_i$ to produce image embeddings. These embeddings are used, in combination with additional context provided by $E_c$, to condition a 3D DDPM. Notably, the pre-trained image encoder $E_i$ can be either CLIP or CISP.
}
    \label{fig:pipeline}
\end{figure*}

\subsection{Joint image-shape embeddings}
Recent advancements in the realm of multimodal learning have been significantly influenced by the development of joint embeddings. This technique involves the projection of data from diverse modalities, such as text and images, into a unified embedding space, enabling enhanced comparison and correlation of features from different data types. Such a unified representation is instrumental in facilitating a more integrated and coherent processing of multimodal information. Joint image-shape embeddings have been investigated by Li \etal~\cite{li2015joint}, who produced a joint embedding space through a multi-step process. First, they construct a shape-only embedding space based on shape similarities, and later learn to pair images with their corresponding shape embeddings. Kuo \etal~\cite{kuo2020mask2cad} adopt a holistic approach, jointly learning image and 3D CAD shape embeddings. This concept is further extended in~\cite{kuo2021patch2cad}, which learns a mapping from image patches to CAD shapes. Imagebind~\cite{imagebind} presents an alternative by aligning multiple modalities (video, audio, depth, termal, IMU) to images using a contrastive approach as in~\cite{CLIP}. However, for 3D data they only focus on depth maps, which contain partial information about the 3D scene, potentially omitting crucial 3D details. We instead incorporate such details in our study.

\section{Building Shape-aware joint embeddings}
\label{sec:joint_embeddings}

To steer the generation of a diffusion model toward samples with particular characteristics, DDPMs are conditioned exploiting a joint embedding space between two modalities. Such space enforces, by construction, that the embeddings of a guiding modality are structured according to the information content of another modality.
Joint embedding spaces are typically constructed using the state-of-the-art contrastive method first described in CLIP~\cite{CLIP} and which has transformed the landscape of multimodal learning.

In the original work, CLIP aligns image and text captioning pairs using a constrastive loss, while also maximizing the distance between non-matching pairs.
The alignment of images to text allows the space to reflect a semantic structure. Nevertheless, this space lacks 3D information, which might be as crucial in downstream tasks involving shape generation.

To build a conditioning embedding space that also encodes semantic and structural shape details,
we introduce a novel contrastive-based joint embedding space aligning image and shape pairs. We refer to this model as CISP (Contrastive Image-Shape Pre-training).
In ways analogous to CLIP, we define an encoder $E_i$ processing images and an encoder $E_s$ processing shapes, both producing embeddings of size $d$. Given a batch containing $N$ \textit{(image, shape)} pairs, our training objective is to maximize the embedding similarity between each image and its corresponding shape, while minimizing the similarity for non-matching pairs.

For each pair in the batch, we compute L2-normalized image embeddings $\mathbf{e_i}$ and shape embeddings $\mathbf{e_s}$,  using $E_i$ and $E_s$ respectively.
Our training loss is then composed of two cross-entropy terms:
\begin{equation}
\small
    \operatorname{L_{CISP}} = \frac{1}{2} \operatorname{L_{i \to s}} + \frac{1}{2} \operatorname{L_{s \to i}},
\end{equation}
with:
\begin{equation}
\small
    \operatorname{L_{i \to s}} = - \frac{1}{N}  \sum\limits_{j=1}^{N} \log{\frac{\exp{\left(\nicefrac{\mathbf{e_i}^j \cdot \mathbf{e_s}^j}{T}\right)}} {\sum\limits_{k=1}^{N}\exp{\left(\nicefrac{\mathbf{e_i}^j \cdot \mathbf{e_s}^k}{T}\right)}}},
\end{equation}
\begin{equation}
\small
    \operatorname{L_{s \to i}} = - \frac{1}{N} \sum\limits_{j=1}^{N} \log{\frac{\exp{\left(\nicefrac{\mathbf{e_s}^j \cdot \mathbf{e_i}^j}{T}\right)}} {\frac{1}{\tau} \sum\limits_{k=1}^{N}\exp{\left(\nicefrac{\mathbf{e_s}^j \cdot \mathbf{e_i}^k}{T}\right)}}},%
\end{equation} 
where $\cdot$ is the inner product, $T=\nicefrac{1}{exp(\tau)}$ with $\tau$ being the temperature parameter, and the apices indicize elements in the batch. 
$\operatorname{L_{i \to s}}$ measures the ability of the model to predict the correct shape given an image, while $\operatorname{L_{s \to i}}$ measures the ability to predict the correct image given a shape. The temperature parameter $\tau$ is used to scale the logits of the softmax and is trained jointly with the network. We initialize the temperature parameter as in~\cite{temperature}, and we clip it following~\cite{CLIP} to prevent training instabilities. This objective function is designed to maximize the similarity of the $N$ matching (image, shape) pairs and minimize the similarity of the $N^2 - N$ unmatching pairs.

As the dataset we use (\cref{sec:experiments}) is of limited size, we employ as image encoder $E_i$ a Data-efficient image Transformer (DeiT)~\cite{DeiT}, i.e., DeiT Base (DeiT-B), using 768-dimensional hidden embeddings, 12 layers with 12 attention heads each and output embeddings of dimension $d=256$.

As shape encoder $E_s$ we adapt the DeiT-B model to a 3D context. We refer to this model as \textit{3D-DeiT}. In particular, we replace the 2D convolutional layers---originally designed for transforming images into patch embeddings---with 3D convolutions for mapping voxel shapes to similar embeddings. This modification provides a significant advantage: we can maintain the remainder of the network architecture identical to the architecture of the image encoder $E_i$.

Consequently, we can initialize the shape encoder $E_s$ with the DeiT-B weights pre-trained on ImageNet~\cite{imagenet}. We observe that this pre-training accelerates the convergence process. We hypothesize that such initialization allows for a rapid alignment of the shape features in the joint space.

To extract the global CISP embedding from the transformers, we substitute the global average pooled query with a learned token prepended to the input sequence, inspired by class tokens first used in NLP~\cite{devlin2018bert} and later introduced in ViT~\cite{Vit}.

\section{3D Diffusion Generation Pipeline}
\label{sec:diffusion}
We exploit the joint embedding space from \ref{sec:joint_embeddings} to condition a DDPM to generate image-aligned 3D shapes. We design our diffusion model to be trained using either CLIP or CISP embeddings indistinguishably. We draw inspiration from well-established text-to-image architectures ~\cite{DALLE2,GLIDE} for their technical implementation.
Our full pipeline is depicted in \Cref{fig:pipeline}. Given an input query image, we obtain its joint image embedding through an image encoder (either from CLIP or CISP), and subsequently project it to multiple tokens via linear learnable layers.
We also implement a separate trainable image encoder $E_c$, as this addition has been shown to help the generation process~\cite{DALLE2}.

For our DDPM model we extend the ADM model~\cite{diffBeatGANs} to function in the 3D domain. 
Specifically: (1) We replace 2D convolutions with 3D convolutions. (2) We use the joint image embeddings from $E_i$ in two ways: first, we project and add them to the timestep embedding; second, in each attention block of the network, we project the joint embeddings into 4 extra tokens and concatenate them to the attention context (keys, values). (3) We prepend 8 learnable tokens to the input of $E_c$ and use the corresponding outputs as additional attention context, as with the joint embeddings. Timesteps are encoded by sinusoidal embeddings~\cite{transformer}. Our DDPM module iteratively refines an input 3D tensor, which is finally transformed into the output shape by binary thresholding. 

To correctly learn to generate conditioned samples, 
we train our CLIP-guided and our CISP-guided model with classifier-free guidance~\cite{cfreeGuidance}. This approach obtains similar results compared to classifier guidance~\cite{ddpm}, while eliminating the need for a separate classifier. To do so, first we jointly train a conditional and unconditional model; then, we make predictions by combining their score estimates to step toward the guidance direction. To jointly train a conditional and an unconditional model, we replace the input conditioning with a learnable null token $\emptyset$ with probability $p$. At inference time, we combine the conditional and unconditional predictions at each step $t$ as:
\begin{equation}
   \medmath{y_\theta(x_t, t|c) = y_\theta(x_t, t|\emptyset) + w\cdot\left(y_\theta(x_t, t|c) - y_\theta(x_t, t|\emptyset) \right)},
\end{equation}
where c is the guidance token(s), $y_\theta$ is the DDPM, $x_t$ is the input at time $t$ of the diffusion process, and $w\geq1$ is the guidance scale. We apply classifier-free guidance with $p=0.1$ on $E_c$ tokens and $E_i$ embeddings independently.

Inference is performed by first generating a pure noise sample, and then running 1000 backward diffusion steps.
We obtain a tensor representing the output volume that we voxelize through binary thresholding. For conditional generation, we apply classifier-free guidance with a guidance scale of $1.5$, as we found it produces the best results.

\section{Experiments}
\label{sec:experiments}
We perform 
experiments to compare and understand 
the effect of guiding the 3D generation process with a text-image joint space (CLIP) versus a 3D-informed joint space (CISP).
We perform two orthogonal quantitative evaluations. 
In \cref{sec:experiments_generation}, we assess the generation capabilities of both models, regardless of their guidance, while in \cref{sec:experiments_coherence} 
we focus explicitly on their coherence to the guidance image.

We further analyze the joint embedding space yielded by CLIP and CISP, investigating its regularity and inherent structure through manifold interpolations (\cref{sec:experiments_interpolation}). 
To further prove the generalization of CISP outside its training dataset, despite it being of smaller size compared to the extensive dataset used for CLIP, we evaluate the out-of-distribution (OOD) generation capabilities of both models with sketched drawings and real-world images (\cref{sec:experiments_ood}). Lastly, we perform ablations on our architecture (\cref{sec:experiments_ablation}).

In all experiments, we focus on the Airplane, Car, and Chair categories from the ShapeNet~\cite{shapeNet} test set, following previous shape generation works~\cite{PVD,luo2021diffusion,lion}. We further stress our models, training them on the Table and Watercraft categories. 

For our CLIP-conditioned DPPM, we rely on the open-source version of CLIP, OpenCLIP~\cite{openclip}, specifically OpenCLIP ViT-B/32 with $d=512$.

\subsection{Generation Capabilities} 
\label{sec:experiments_generation}
We undertake a comparative analysis of the shape generation quality and diversity between our CISP-driven model and the CLIP-driven model. 
Quality is defined as the fidelity of the generated shapes, without accounting for their alignment with the reference image or their variation. Diversity, instead, pertains to the range of structural variations in shapes, disregarding their logical structure or alignment with the input image. Both measurements are crucial for a well-performing generative model.

Our comparison focuses primarily on the two presented models; however, for context, we also reference the performance of recent 3D Deep Diffusion Probabilistic Models~\cite{wavelet,luo2021diffusion,PVD,lion}.
Given that quality and diversity evaluations in existing literature are typically unguided, we adjust our methodology accordingly to ensure a fair and equivalent comparison. Our approach to unconditioned generation involves removing guiding images in our model and replacing $E_i$ embeddings and $E_c$
tokens with pre-learned null tokens. The primary metric for our comparative analysis is the 1-Nearest Neighbor Accuracy (1-NNA), as it is acknowledged as the most indicative metric for 3D generation~\cite{PointFlow}, addressing issues of other metrics such as Matching Distance (MMD) and Coverage (COV).

1\nobreakdash-NNA, introduced in~\cite{1nna} and later applied to 3D generation in~\cite{PointFlow}, evaluates the diversity and quality of generated samples through a 1-Nearest Neighbor classifier's accuracy. We define $S_g$ as the set of generated samples and $S_r$ as the reference samples set, with $|S_r|=|S_g|$. For a sample $x$, let $N_x$ represent its nearest neighbor, where ${N_x \in \{S_g \cup S_r - x\}}$. 1\nobreakdash-NNA is then computed as:
\begin{equation}
    \medmath{\operatorname{1-NNA}(S_g, S_r) = \frac{\sum\limits_{x \in S_g} \mathbb{I}[N_x \in S_g]+\sum\limits_{x \in S_r} \mathbb{I}[N_x \in S_r]}{|S_g|+|S_r|}}.
\end{equation}

Here, $\mathbb{I}$ is the indicator function. The ideal 1\nobreakdash-NNA score is $50\%$, indicating a precise classification of generated and reference shapes by the 1\nobreakdash-NN classifier.

To adapt voxel shapes for EMD and CD analysis, we follow the procedure outlined in PVD~\cite{PVD}, sampling 2048 points from each generated shape's surface. Our methodology also incorporates metrics implementations from PVD's publicly available code. 

\begin{table}[t]
    \renewcommand{\arraystretch}{1.25} 
    \caption{Generation capabilities of CLIP- and CISP-guided DDPMs\\
    \scriptsize{Literature 3D DDPM models are reported as a reference}}
    \label{tab:1nnaresults}
    \begin{adjustbox}{width=0.9\linewidth,center} 
    \begin{tabular}{ll@{\hskip 47pt}c@{\hskip 17pt}c@{}} 
        \toprule 
        \multicolumn{4}{c}{1\nobreakdash-NNA}\\
        \midrule
        Shape & Model & CD & EMD \\ 
        \midrule
        \multirow{6}{*}{Aeroplane} & \cite{wavelet} & 71.69 & 66.74 \\
        & \cite{luo2021diffusion} & 62.71 & 67.14 \\
        & PVD~\cite{PVD} & 73.82 & 64.81 \\ 
        & LION~\cite{lion} & 67.41 & 61.23 \\ 
        \cmidrule{2-4} 
        & Ours (CLIP) & 63.37 & 59.79 \\ 
        & Ours (CISP) & \underline{\textbf{58.93}} & \underline{\textbf{56.93}} \\ 
        \bottomrule
        \multirow{6}{*}{Car} & \cite{wavelet} & - & - \\ 
        & \cite{luo2021diffusion} & - & - \\
        & PVD~\cite{PVD} & 54.55 & 53.83 \\ 
        & LION~\cite{lion} & 53.70 & \underline{52.34} \\ 
        \cmidrule{2-4}
        & Ours (CLIP) & \underline{\textbf{53.17}} & 53.97 \\
        & Ours (CISP) & 53.20 & \textbf{53.11} \\ 
        \bottomrule
        \multirow{6}{*}{Chair} & \cite{wavelet} & 61.47 & 61.62 \\ 
        & \cite{luo2021diffusion} & 62.08 & 64.45 \\ 
        & PVD~\cite{PVD} & 56.26 & 53.32 \\ 
        & LION~\cite{lion} & 53.41 & \underline{51.14} \\ 
        \cmidrule{2-4}
        & Ours (CLIP) & 53.52 & \textbf{51.69} \\ 
        & Ours (CISP) & \underline{\textbf{53.30}} & 51.97 \\ 
        \bottomrule
        \multicolumn{4}{l}{$^{\mathrm{a}}$\scriptsize{Best of our models highlighted in bold.}} \\
        \multicolumn{4}{l}{$^{\mathrm{b}}$\scriptsize{Best overall model is underlined.}}
    \end{tabular}
    \end{adjustbox}
\end{table}

The results, detailed in \Cref{tab:1nnaresults}, show that both our models obtain high quality and diversity, comparable or superior to the literature in this domain. When compared, CLIP- and CISP-guided models achieve similar quality and diversity across most categories. An exception is noted in the Aeroplane category, where the CISP-guided model shows notable improvements of $7.5\%$ and $5\%$ in CD and EMD metrics, respectively. Although similar scores were anticipated for both models due to the unguided nature of the experiments, which does not account for coherence, the CISP-guided model's training involved learning from a 3D-informed space. This aspect likely contributed to its enhanced performance in certain categories, such as aeroplanes, indicating a better grasp of 3D semantics and structural features.

\subsection{Image Coherence}
\label{sec:experiments_coherence}

\begin{table}[t]
    \renewcommand{\arraystretch}{1.25}
    \caption{Coherence of CLIP- and CISP-guided DDPMs}
    \label{tab:coherence}
    \begin{adjustbox}{width=0.8\linewidth,center}
    \begin{tabular}{ll|cccc} 
    \toprule 

    \multicolumn{2}{r}{$\#$ of samples} & 1 & 5 & 10 & 15 \\ 
    \hline
    \multirow{2}{*}{IoU}&CLIP-Guided & 0.537 & 0.573 & 0.586 & 0.633 \\ 
    & CISP-Guided & \textbf{0.579} & \textbf{0.633} & \textbf{0.649} & \textbf{0.658} \\ \midrule

    \multirow{2}{*}{F-Score}&CLIP-Guided & 0.298 & 0.321 & 0.333 & 0.340 \\ 
    &CISP-Guided & \textbf{0.363} & \textbf{0.402} & \textbf{0.414} & \textbf{0.421} \\ \bottomrule
    \multicolumn{5}{l}{$^{\mathrm{a}}$\scriptsize{The best model is highlighted in bold.}}
    \end{tabular}
    \end{adjustbox}
\end{table}

\begin{figure}[t]
  \centering
   \includegraphics[width=0.9\linewidth]{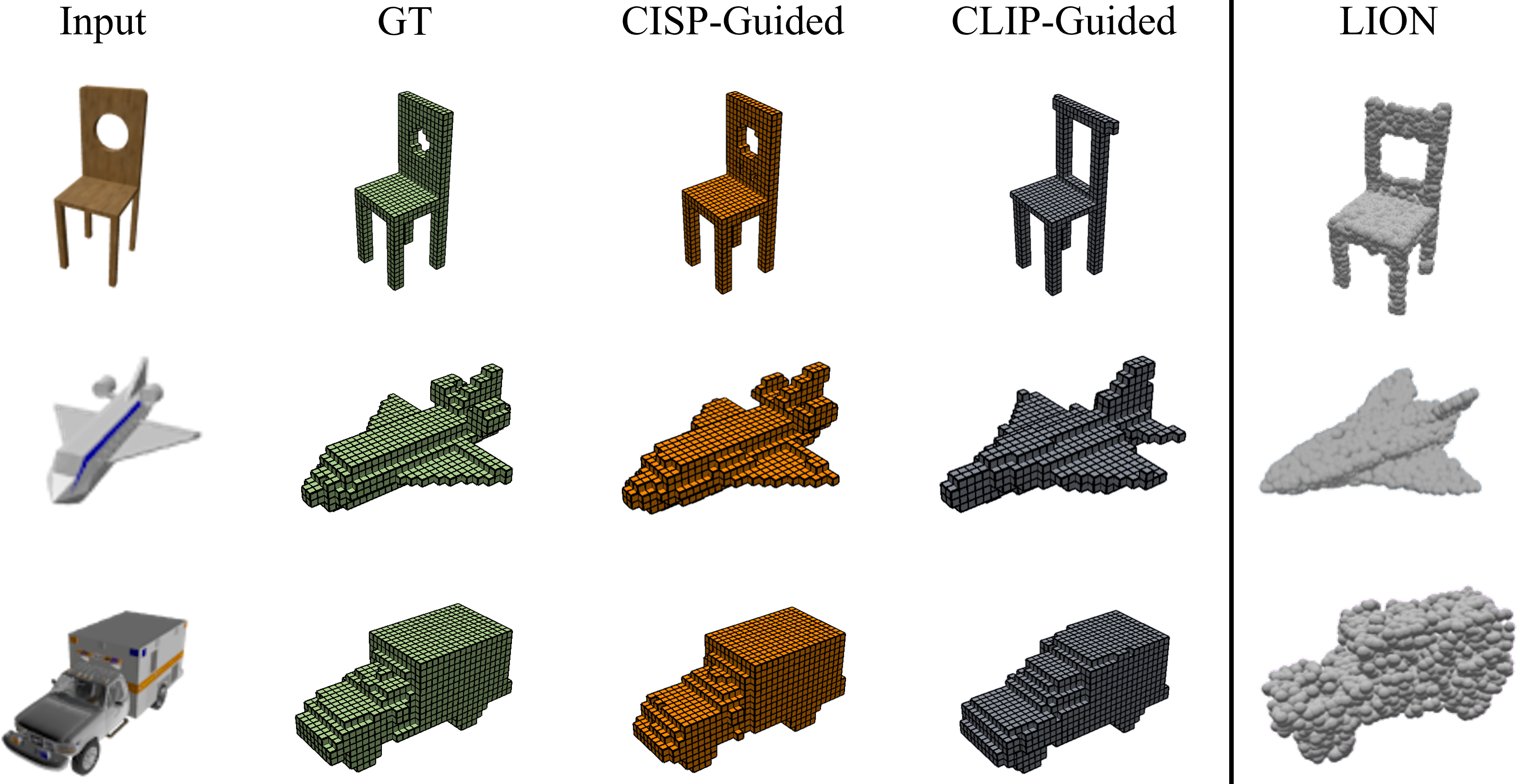}
   \caption{Examples of image-guided shape generation with CISP-guided and CLIP-guided models. We also report a point cloud generation from LION~\cite{lion}, also guided with CLIP. Notice how all CLIP-guided models are biased towards the same structural mistakes (e.g., chair backrest hole, airplane tail engines).}
    \label{fig:comparison}
\end{figure}

\begin{figure*}[t]
  \centering
   \includegraphics[width=0.9\linewidth]{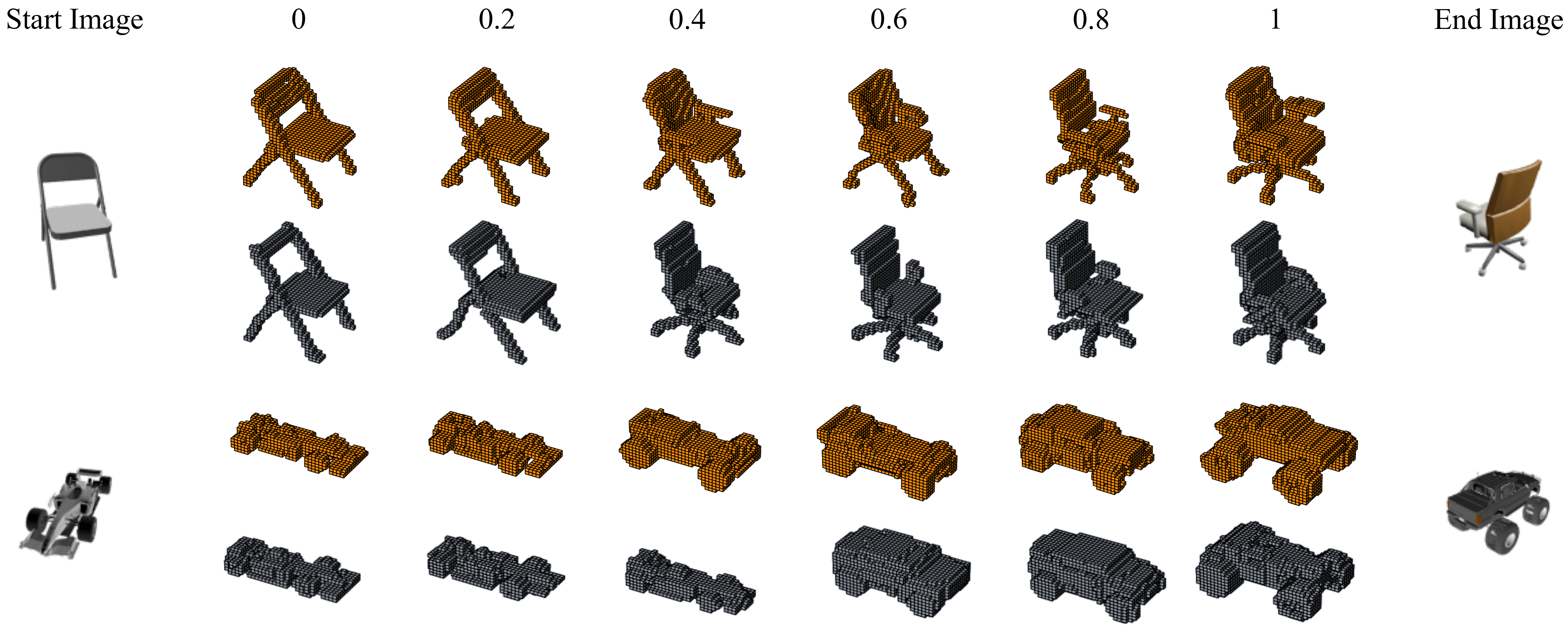}
   \caption{We interpolate embeddings between Start and End Images and generate shapes with our CISP-guided DDPM (orange) and CLIP-guided DDPM (grey). The CISP-guided model displays a smoother transition in terms of structural 3D components. The chair in orange slowly grows wheels and armrests and mutates its backrest, as opposed to a sharp style change in the chair in gray. Similarly, the orange racecar slowly changes height, wheel size, and overall shape to become a monster truck, in contrast with the abrupt change seen in gray.}
    \label{fig:interpolation}
\end{figure*}

Having established the proficiency of both models in generating shapes of high quality and diversity, we advance to evaluating their coherence. Coherence is defined as the structural congruence with the desired 3D object depicted in the reference image. We notice that although coherence may imply quality, the converse is not necessarily true.

This analysis utilizes metrics from the domain of 3D reconstruction to measure how closely the shapes produced by the models align with the ground truth shapes depicted in the guiding images. Coherence, in this sense, refers to the degree to which the generated shape preserves the structural and geometric characteristics of the target shape. Reconstruction metrics allow for the evaluation of various aspects of the generated shapes, including geometric fidelity, topological accuracy, surface quality, and detail preservation. We employ two primary metrics for this assessment: Intersection over Union (IoU) and F-Score. The IoU metric is voxel-based and quantifies the overlap between the generated shape and the ground truth by calculating the ratio of their intersection to their union. On the other hand, the F-Score, as proposed in \cite{whatdo3dlearn}, is derived from point cloud data and provides a more granular measurement. It is especially useful in scenarios where IoU may not adequately capture the nuanced geometric details of the generated shapes. We compute the F-Score by sampling 2048 points from both the predicted and the target shapes, and then calculating the F-Score@1\% as detailed in~\cite{whatdo3dlearn}.

To account for the inherent variability in our generative models, we conduct evaluations across a progressively larger number of generated shapes for each input sample, recording the maximum metric value obtained for each set. This approach allows us to capture the best possible performance of the models under stochastic conditions. The detailed outcomes of these evaluations are presented in \cref{tab:coherence}, where it is evident that the CISP-guided model consistently outperforms the CLIP-guided model in terms of structural coherence with the ground truth.

A closer examination of the coherence of both models is facilitated by the qualitative comparison in \cref{fig:comparison}.
The CLIP-guided model demonstrates a commendable ability to grasp and replicate the general structure of objects. However, it falters when replicating the finer structural nuances within these categories. Such disparity becomes apparent in detailed features such as the aircraft's tail and engines, or the circular cutouts in the chairs. These observations, confirmed by additional examples presented in later sections (\cref{sec:experiments_sketch2shape,sec:experiments_inthewild}), highlight the finer granularity at which the CISP-guided model represents 3D objects.

Interestingly, a similar pattern of structural inaccuracies is observed in the LION model \cite{lion}, which also utilizes CLIP guidance.
This similarity highlights the common challenges faced when non-3D informed embeddings guide the generative process. 

\begin{figure*}[t]
  \centering
   \includegraphics[width=0.9\linewidth]{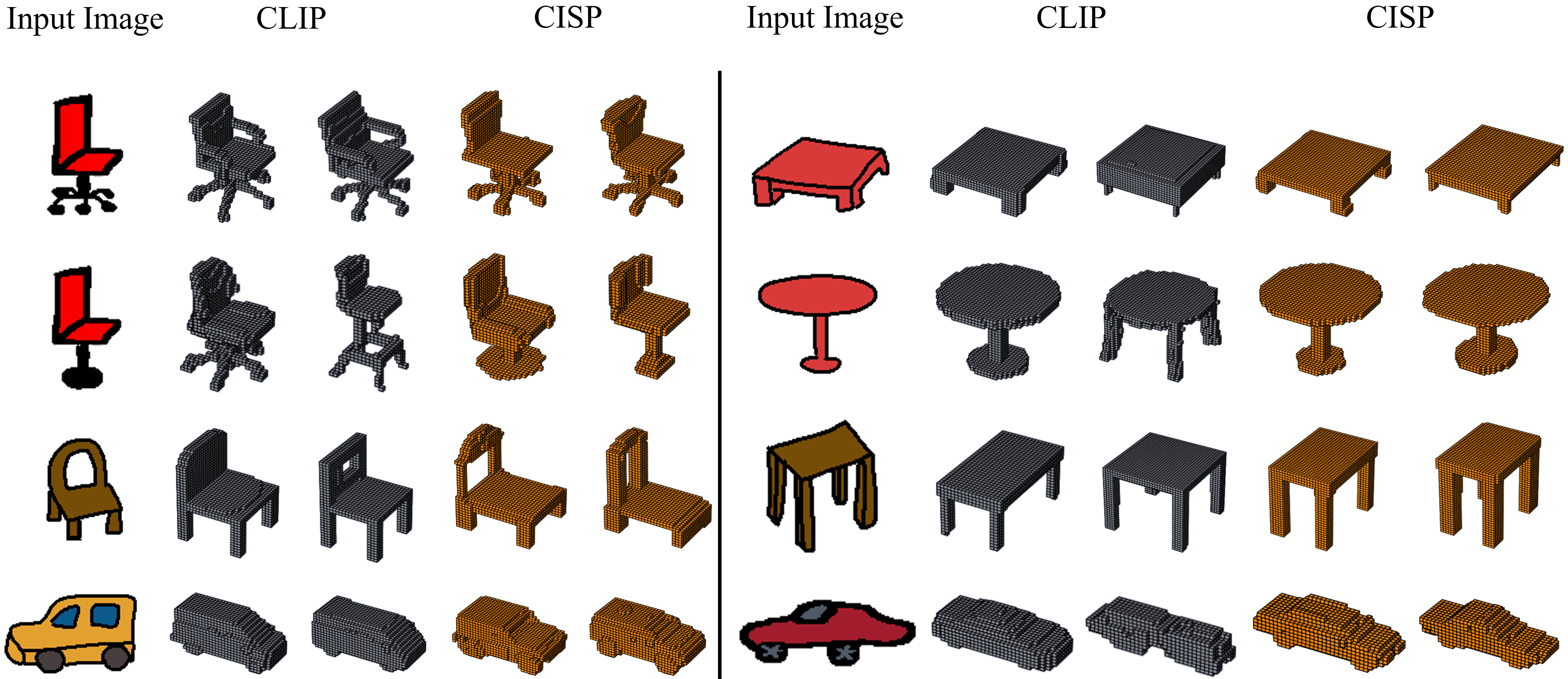}
   \caption{Generation results from hand-drawn sketches, proving both models' generalization capabilities and highlighting CISP's higher attention to structural details.}
    \label{fig:sketch2shape}
\end{figure*}
\subsection{Embedding Space Interpolation}
\label{sec:experiments_interpolation}

We compare the regularity of CLIP and CISP joint embedding spaces and their impact on the generated shapes through latent manifold interpolations. We start from a pair of images $A$ and $B$, representing objects with different structural features and details. With each model, we then generate the image joint embeddings $\mathbf{e}^A, \mathbf{e}^B$, and interpolate between these values in the joint embedding space. At each interpolation step, we generate a new shape via diffusion.
We employ spherical linear interpolation (Slerp)~\cite{slerp}, which we have determined to yield superior results for both models compared to standard linear interpolation. SLERP is defined as:
\begin{equation}
    \operatorname{SLERP}(\mathbf{e}^A, \mathbf{e}^B, \alpha) = \frac{\sin{\left((1-\alpha)\, \theta\right)}}{\sin{\theta}}\mathbf{e}^A+\frac{\sin{\left(\alpha\, \theta\right)}}{\sin{\theta}}\mathbf{e}^B,
\end{equation}
where $\theta$ is the angle between $\mathbf{e}^A$ and $\mathbf{e}^B$.
We vary the interpolation factor $\alpha$ from 0 to 1, with increments of 0.2 for each sample, for a total of 6 interpolation steps.

A qualitative comparison of the two pipelines' interpolation capabilities is documented in \cref{fig:interpolation}.
We find that both models create structurally realistic shapes, proving the smoothness of their guiding embedding spaces. However, it is noteworthy that the transitions in the shapes generated by the CLIP-guided model are considerably more abrupt, often resulting in a stark metamorphosis from the initial to the final form. In contrast, the CISP-guided model exhibits more gradual transitions, incorporating structural elements of both source and target shapes in a progressive manner. For instance, in the first row, we observe how the original chair structure gradually acquires wheels and armrests before its final transformation into an office chair. Analogously, the Formula 1 car progressively contracts in length, increases in height and enlarges its wheels as it morphs into a monster truck.

The nuanced transitions and the preservation of characteristic features during the interpolation process with the CISP-guided model underscore its deeper understanding of spatial relationships and structural semantics in three-dimensional objects. This is in stark contrast with the CLIP-guided model, which, despite showing robustness in this domain, appears to lack the same degree of familiarity with three-dimensional representations.

\subsection{OOD Generalization}
\label{sec:experiments_ood}
In the scope of this study, we train both diffusion models to generate shapes from the subset of ShapeNet standardly used in the generation literature~\cite{PVD,luo2021diffusion,lion}. We use this same dataset also to pre-train CISP. This, however, is a small-scale dataset when compared to the amount of data used for pre-training CLIP. We expect the CLIP-conditioned DDPM to benefit from this fact.

To verify the robustness of both pipelines and the impact of the different pre-training scale, we study their qualitative performance on out-of-distribution (OOD) samples coming from hand-drawn sketches and real-world images.
While we expect the CLIP-based pipeline to handle these samples with ease, we question whether the CISP-based model would handle shifted data distributions well, and prove that its higher coherence is not attributed to overfitting the original shape dataset.

\subsubsection{Sketch to Shape}
\label{sec:experiments_sketch2shape}
\hfill 

We evaluate the shape generation using hand-drawn sketches as guiding images.
We operate in a zero-shot manner, i.e., without any fine-tuning on the novel data. The sketches, which are entirely hand-drawn, represent common objects, including various types of chairs, tables, and vehicles. 
We report samples from this sketch-to-shape generation using the CLIP- and CISP-guided models in~\Cref{fig:sketch2shape}. 

We find that both models generalize well to hand-drawn sketches, producing structurally sound and realistic shapes, even when fed with simplistic and under-detailed images. While we expected this behavior from the CLIP-based pipeline, it is interesting how even the small-scale CISP pipeline correctly interprets simple drawings.
Furthermore, we qualitatively notice that, even in this scenario, the CISP-based generation achieves higher visual coherency to the input image. Indeed, while the CLIP-guided model captures the general form and structure of the sketched objects, it struggles to produce the level of detail and structural fidelity provided by CISP-conditioning.
This is particularly evident for chairs, where the CISP-based model closely reproduces the sketched armrests and wheels.

\subsubsection{In-the-wild capabilities}
\label{sec:experiments_inthewild}
\hfill 

\begin{figure}[t]
  \centering
   \includegraphics[width=0.9\linewidth]{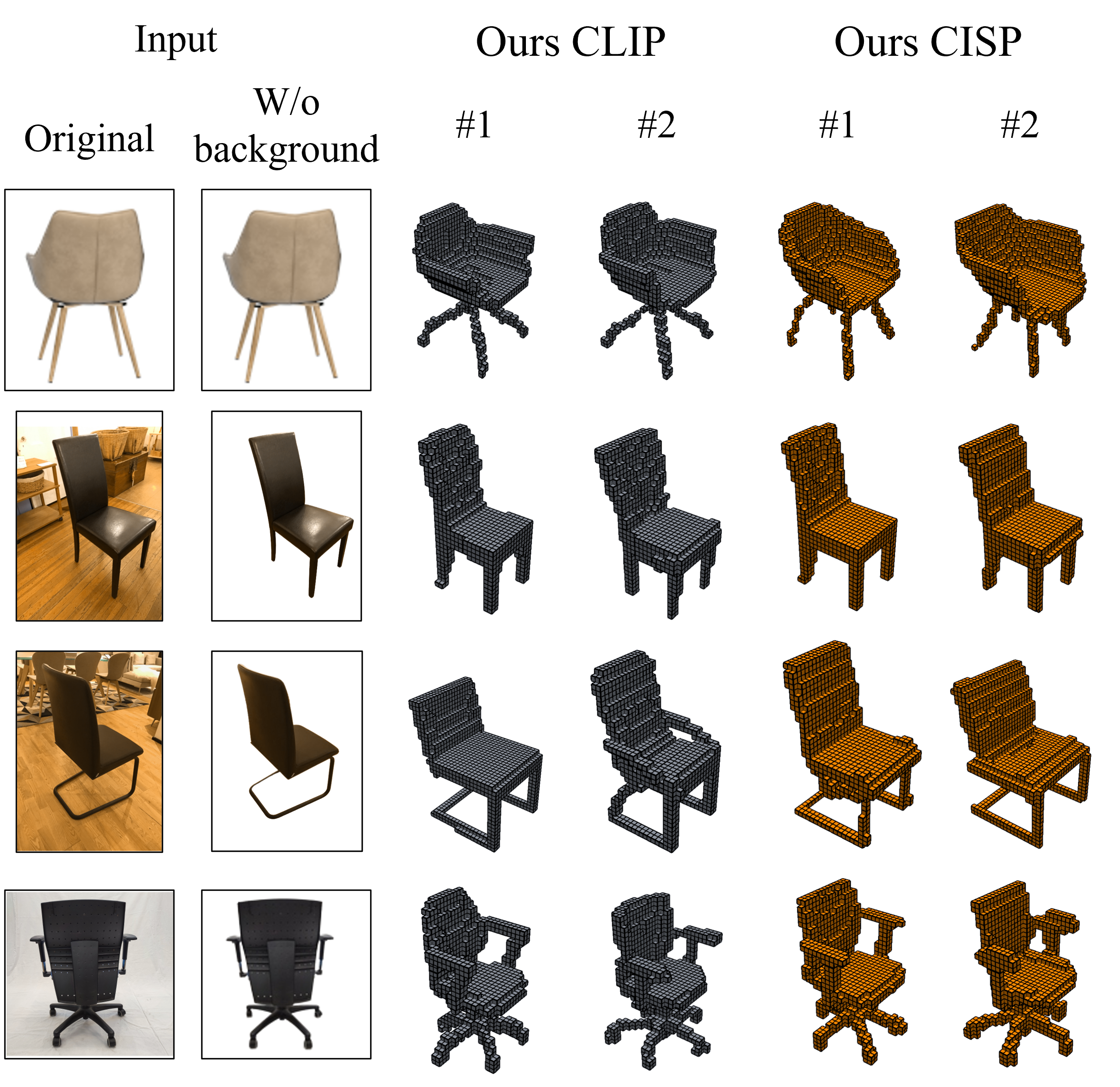}
    \caption{Generation results from real-world images. Top-to-bottom: chair from an online catalog with frontal occlusion; Pix3D~\cite{pix3d} real-world chairs seen from front and back; smartphone photograph of the authors' office chair.}
    \label{fig:iiw}
\end{figure}

We present several examples of in-the-wild image-conditioned generation in \cref{fig:iiw}. We demonstrate conditioning from three increasingly hard in-the-wild sources: (1) an online chair catalog, (2) a well-known real-world image-shape dataset (Pix3D~\cite{pix3d}), and (3) photographs of an office chair taken by the authors using a smartphone. 
In this regard, it is important to make a specification. 
Our training dataset, ShapeNet, contains only images with a blank background.
For this reason, CISP is not capable of processing as-is images with generic backgrounds. Nevertheless,  we find that an automated background removal method effectively resolves this limitation and provides the model with an image in the expected format.
Conversely, CLIP, benefitting from its comprehensive pre-training, can generally handle images with any background. Nonetheless, we observe a decrease in the generation quality of the CLIP-based pipeline when using backgrounds, which we deem due to the DDPM not being trained on OOD CLIP embeddings containing information about the background. 
For this reason, when guiding any of our models with in-the-wild images, we always first perform an automated background removal. This simple addition, completely automated, leads both models to generate shapes consistent with a real-world guiding image, even in the presence of realistic lighting effects and occlusions. 
As in the previous case, the CISP-conditioned shapes display more accurate details. This is evident, for instance, when looking at the locations where the chair legs are attached to the main body. In CISP, they respect the geometry of the conditioning image, even if such configuration is uncommon in typical chairs. The CLIP-based model, instead, is more prone to generate common chair features, even if it means not adhering to its guidance.

\subsection{Ablation study}
\label{sec:experiments_ablation}
We conduct an ablation study on the architecture and embedding dimension used in 3D-DeiT, the novel shape encoder introduced with CISP. In particular, we studied different configurations of a transformer model and a convolutional neural network (CNN).
The transformer architecture adheres to the structure described in~\cref{sec:diffusion}. 
For the CNN architecture, instead, we used the downsampling component of the ADM model proposed by Dhariwal and Nichol~\cite{diffBeatGANs}, which corresponds to a UNet~\cite{UNet} equipped with global attention at lower resolution layers. In this case, features are progressively downsampled to a vector and subsequently projected to the desired embedding dimension via a convolution with a kernel size of 1.
For each architecture, we compare results using different embedding dimensions.

We assess both architectures on Top-k accuracy in a retrieval setting. A batch (of size 128) of paired images and shapes is randomly sampled, embedded through our model, and a similarity matrix is computed for all shape and image embedding pairs.
For each image embedding, we retrieve the most similar $k$ shape embeddings in the batch and evaluate whether the corresponding shape is within them. The top-k accuracy evaluates the percentage of images for which a correct match is found. An analogous value is computed retrieving images from shape embeddings, and the two metrics are averaged.

We compute this metric for $k = 1,\dots ,5$ in order to account for scenarios where the batch contains very similar shapes. In ShapeNet~\cite{shapeNet}, indeed, it is quite common to encounter objects significantly hard to distinguish even for the human eye. This is especially true for untextured 3D data.

\Cref{tab:ablation} reports the Top-k accuracy for each considered configuration.
Our main finding is twofold. First, transformers consistently outperform the corresponding CNNs, regardless of embedding dimension and batch size.
Second, increasing the training batch size has a more positive effect than increasing the embedding dimension. Indeed, using a larger batch size increases the likelihood of finding similar objects in the batch. As a consequence, the model is forced to solve a more difficult task and thus learn more fine-grained representations to correctly match image-shape pairs. 
Lastly, we trained transformer configurations also with a batch size of 64. We did not perform the same experiment using a CNN, since this batch size dimension would not have been feasible with our computation budget. Nevertheless, given the observed trend, we expect it not to exceed the transformer's performance.
The best-performing model, marked in bold in \cref{tab:ablation}, is used in the rest of this work.

\begin{table}[t]
    \renewcommand{\arraystretch}{1.25} 
    \vspace{.5em}
    \caption{Top-k accuracy comparison for different configurations of $E_s$\\\scriptsize{C = ADM CNN + Attention, T = 3D-DeiT}}
    \label{tab:ablation}
    \centering
    \begin{tabular}{ccc|ccccc} 
    \toprule 
        Model & $N$ & $d$ & Top-1 & Top-2 & Top-3 & Top-4 & Top-5 \\ \hline
        C & 16 & 192 & 0.566 & 0.761 & 0.845 & 0.890 & 0.926 \\
        C & 16 & 256 & 0.575 & 0.773 & 0.847 & 0.901 & 0.920 \\ 
        C & 32 & 192 & 0.619 & 0.781 & 0.874 & 0.944 & 0.941 \\
        C & 32 & 256 & 0.639 & 0.792 & 0.868 & 0.910 & 0.949 \\ \hline
        T & 32 & 192 & 0.647 & 0.822 & 0.899 & 0.932 & 0.955 \\ 
        T & 32 & 256 & 0.668 & 0.854 & 0.895 & 0.948 & 0.970 \\
        T & 64 & 192 & 0.695 & 0.864 & 0.924 & 0.957 & 0.980 \\ 
        \textbf{T} & \textbf{64} & \textbf{256} & \textbf{0.709} & \textbf{0.871} & \textbf{0.938} & \textbf{0.963} & \textbf{0.981} \\ \bottomrule
        \multicolumn{8}{l}{$^{\mathrm{a}}$\scriptsize{The best model is highlighted in bold.}}
    \end{tabular}
\end{table}

\section{Discussion}
\label{sec:discussion}
In this study, we thoroughly evaluated the impact of integrating 3D data into the conditioning process for generating 3D shapes using DDPMs, comparing this approach with CLIP guidance. We introduced the Contrastive Image-Shape Pre-training (CISP) model, tailored to align 2D images and 3D shapes within a unified embedding space. We designed an image-conditioned shape generation pipeline exploiting a joint embedding space, which we used to train distinct models, one conditioned on the CLIP embeddings and the other on the CISP embeddings. We then evaluated and compared the 3D shape generation abilities of the two models.

Our results affirm that CISP achieves comparable generation quality and diversity to CLIP, while significantly enhancing the coherence of generated shapes with input images. The embedding space of CISP, enriched with 3D structural knowledge, contributed to this heightened coherence. This suggests that a deeper understanding of the 3D domain can indeed be leveraged to improve the generative quality of models in terms of structural and semantic accuracy.

Moreover, our exploration into out-of-distribution (OOD) generalization revealed that CISP, despite being trained on a smaller-scale dataset compared to CLIP, exhibited robust performance. This robustness is critical, indicating that CISP's superior coherence is not just a result of overfitting but rather its intrinsic ability to apprehend and generalize 3D semantic subtleties across varied scenarios.

Our study's implications extend beyond the immediate results. Firstly, the success of CISP in a relatively smaller-scale setting suggests promising avenues for scaling up such models. Investing in large-scale multimodal systems that incorporate explicit 3D representations could pave the way for significant advancements in the field of 3D visual content synthesis. Secondly, the methodology and findings of this research contribute to the ongoing dialogue on the role of modality-specific information in multimodal learning systems.
{\small
\bibliographystyle{IEEEtran}
\bibliography{IEEEabrv, references}
}

\end{document}